\documentclass{article}

\usepackage{microtype}
\usepackage{graphicx, float}
\usepackage{booktabs} 
\usepackage{multirow}
\usepackage{makecell}

\usepackage{hyperref}

\usepackage[accepted]{icml2023}

\usepackage{amsmath}
\usepackage{amssymb}
\usepackage{mathtools}
\usepackage{amsthm}
\usepackage{subcaption}

\usepackage[capitalize,noabbrev]{cleveref}

\theoremstyle{plain}

\theoremstyle{definition}

\theoremstyle{remark}

\usepackage[textsize=tiny]{todonotes}

\usepackage{pgfplots}
\pgfplotsset{width=10cm,compat=1.9} % set the width of the plot and the compatibility mode

\icmltitlerunning{What’s left can’t be right - The remaining positional incompetence of contrastive
vision-language models}

\begin{document}

\twocolumn[
\icmltitle{What's left can't be right - The remaining positional incompetence of contrastive vision-language models}

\icmlsetsymbol{equal}{*}

\begin{icmlauthorlist}
\icmlauthor{Nils Hoehing}{ucd,yyy}
\icmlauthor{Ellen Rushe}{trinity}
\icmlauthor{Anthony Ventresque}{trinity}
\end{icmlauthorlist}

\icmlaffiliation{yyy}{SFI Centre for Research Training in Machine Learning at University College Dublin, Ireland}
\icmlaffiliation{ucd}{School of Computer Science, University College Dublin, Ireland}
\icmlaffiliation{trinity}{School of Computer Science and Statistics, Trinity College Dublin, Ireland}

\icmlcorrespondingauthor{Nils Hoehing}{nils.hohing@ucdconnect.ie}

\icmlkeywords{compositionality, vision-language models, contrastive learning, clip, visual genome relations}

\vskip 0.3in
]



\printAffiliationsAndNotice{}
% leave blank if no need to mention equal contribution

\begin{abstract}
Contrastive vision-language models like CLIP
have been found to lack spatial understanding capabilities. In this paper we discuss the possible causes of this phenomenon by analysing both datasets and embedding space. By focusing on simple left-right positional relations, we show that this behaviour is entirely predictable, even with large-scale datasets, demonstrate that these relations can be taught using synthetic data and show that this approach can generalise well to natural images - improving the performance on left-right relations on Visual Genome Relations.
\end{abstract}
\section{Introduction}
\label{introduction}

While contrastive vision-language models succeed on traditional image-text retrieval tasks like COCO~\cite{coco} and Flickr30k~\cite{flickr30k}, they still lack some forms of compositional understanding. Compositionality refers to the way words can be combined to form meaning. In vision-language models this is expressed as a combination of visual content. Visual understanding is a task that underpins compositionality since, if the meaning of words or visual content is truely ``understood", any combination of different content should still be correctly interpreted, regardless of how common that combination is in the real world. Models that simply memorize the most common combinations of content cannot be said to truely ``understand" this content since it will not be able to correctly interpret new combinations.

One aspect of compositionality that both generative and discriminative vision-language models struggle to achieve is relative spatial understanding. In CLIP's~\cite{clip} embedding space ``A horse to the left of a tree" has almost the same representation as ``A horse to the right of a tree". Stable Diffusion~\cite{stablediffusion} and DALL-E 2~\cite{dalle2} also do not reliably generate objects in the correct positions \cite{dall-eval, preliminarydalle, benchspat}. 

With the aim of addressing these deficiencies in spatial understanding, we attempt to explain the root causes of failed positional understanding in contrastive vision-language models and reveal why these failures cannot be remedied by simply increasing the scale of datasets and models arbitrarily. We show how  the distances between representations in CLIP's (and CLIP variants') embedding space are unaligned with the distances their true meaning would imply. Finally, using the specific case of left-right positional understanding, we show that we can substantially improve the accuracy on this task for the Visual Genome (VG) Relations benchmark through the use of carefully constructed contrastive pairs. We also show that this improvement leads to an alignment of left-right relations in embedding space that is more representative of their true semantic differences.

The remainder of the paper is organised as follows: Section \ref{related-work} discusses the existing benchmarks used by contrastive vision-language models; Section \ref{background} gives some background on relative position understanding, shows how relative position is encoded in embedding space and the importance assigned to relative position in current visual-language models; In Section \ref{experimental-setup}, we propose a method of learning relative position more accurately using automatically generated contrastive pairs. Section \ref{results} shows the performance of our proposed method on the VG-Relations benchmark and a discussion on the impact of this method and its limitations; Finally we conclude and discuss some possible future work in Section \ref{conclusion}.

\section{Related Work}
\label{related-work}

A large set of benchmarks has been established for evaluation of spatial understanding in vision-language models. Synthetic datasets like NLVR \cite{nlvr}, CLEVR \cite{clevr} and TraVLR \cite{travlr} test for visio-linguistic spatial understanding with simple shapes framed as entailment and visual question answering tasks, though it is not clear to what extent results on these datasets may transfer to natural images. There are also several single modality benchmarks. SpartQA \cite{spartqa}, for instance, is a question answering benchmark for spatial understanding for language-only models. CVR \cite{cvr} is a purely visual test for spatial reasoning that requires models to select the outlier among a group of images. \citet{drawbench}, \citet{relreasoningbenchmark} and \citet{dall-eval} evaluate text-to-image performance and also include some tests for spatial understanding. \citet{benchspat} develop the VISOR benchmark for spatial understanding in text-to-image models and evaluate models by using object-detection. All of them find rather poor spatial understanding in all popular image generation models.
Visual relationship detection is commonly based on object detectors \cite{vrrvg, unifiedvisrel} since these have an inherent understanding of space. This however limits the models to a fixed set of pretrained detectable objects.

Contrastive vision-language models like CLIP \cite{clip} have been found to produce strong visio-linguistic representations that are helpful in all kinds of downstream tasks \cite{clip-benefit}, but their performance on spatial understanding tasks is very poor. \citet{vsr}, for example, find that while all models tested on their Visual Spatial Reasoning benchmark struggle, object detector based LXMERT \cite{lxmert} and ViLT \cite{vilt} are roughly 10 percent stronger than CLIP --likely due to the fact that they explicitly encode positional information.

\section{Relative Position Understanding}
\label{background}
\begin{figure}[h!]
\centering
  \begin{subfigure}{0.49\linewidth}
    \includegraphics[width=\linewidth]{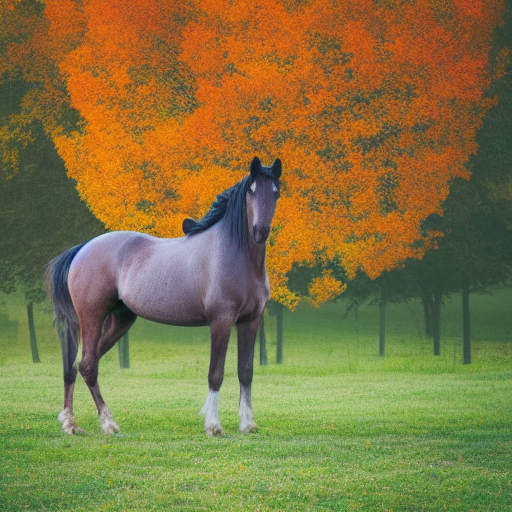}
  \end{subfigure}
  \hfill
  \begin{subfigure}{0.49\linewidth}
    \includegraphics[width=\linewidth]{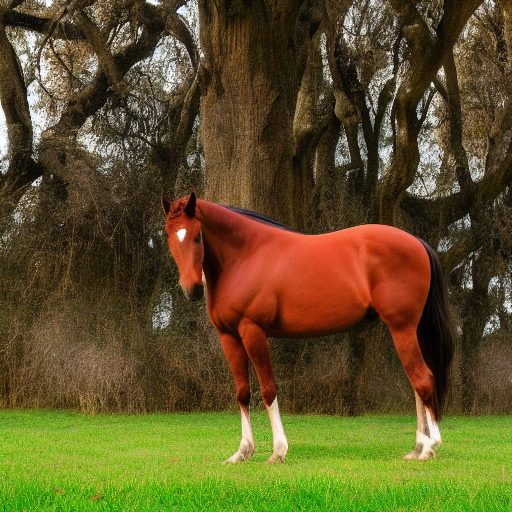}
  \end{subfigure}
  \vspace{0.5cm}
  \begin{subfigure}{0.49\linewidth}
    \includegraphics[width=\linewidth]{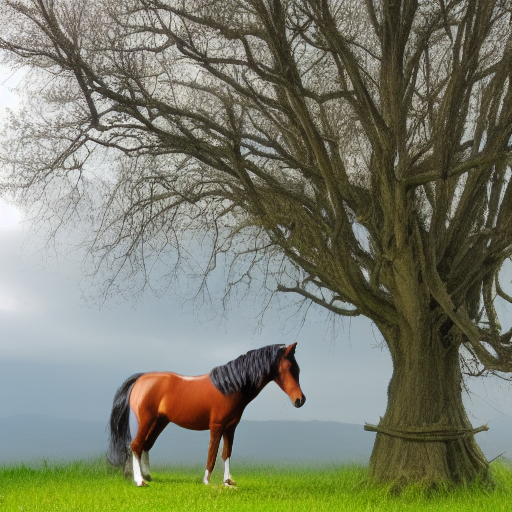}
  \end{subfigure}
  \hfill
  \begin{subfigure}{0.49\linewidth}
    \includegraphics[width=\linewidth]{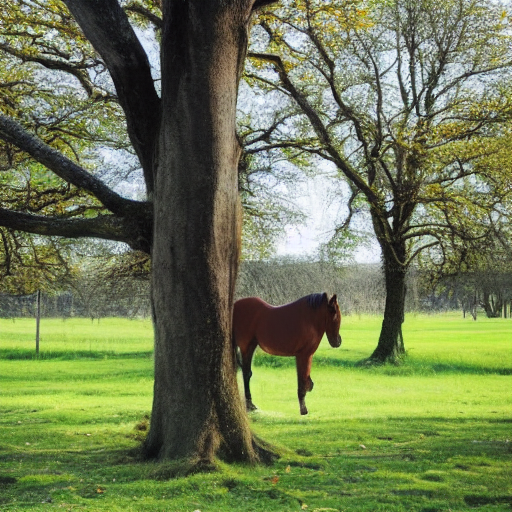}
  \end{subfigure}
  \caption{Stable Diffusion 1.4 generations for the prompt ``An image of a horse standing to the right of a tree"}
  \label{fig:stablediff}
\end{figure}

A fundamental aspect of visual understanding is the concept of relative position understanding, i.e. the position of one particular object or person relative to another. Relative position can encompass a number of concepts, such as ``top", ``bottom", ``left" and ``right".  In this work, we argue that relative position understanding is a key aspect of compositionality, as a model that cannot correctly encode relative position will induce inaccuracies in downstream tasks, such as incorrect positioning of objects and people in generated images. Several recent works have explored the ability of text-to-image models to encode positionality.   \citet{testingrel} and \citet{preliminarydalle} showed that the text-to-image model DALL-E 2 \cite{dalle2}, fails to capture some aspects of compositionality, positioning included. This too can also be observed in other image generators like Stable Diffusion \cite{stablediffusion}.   Figure \ref{fig:stablediff} shows that Stable Diffusion fails to generate images with objects in the correct positions -  even for very simple scenes composed only of two objects/people and a relation. Several solutions have been proposed \cite{composablediff, spatext, controlnet, gligen, directeddiff}, but all rely on additional inputs such as bounding boxes that ground the location of the objects to specific coordinates in the image. This indicates that no actual understanding of the positional information specified in the text has been achieved.

\begin{figure}[h!]
    \includegraphics[width=\linewidth]{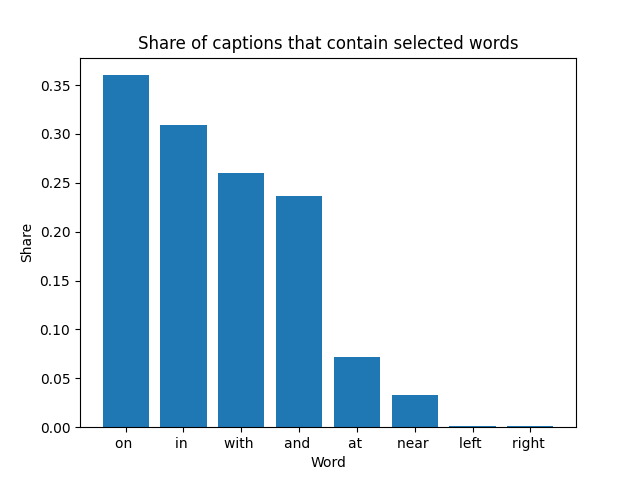}
    \caption{Proportion of COCO captions that contain selected words}
  \label{fig:word_distribution}
\end{figure}

The same deficiency likely exists in vision-language models like CLIP \cite{clip}, BLIP \cite{blip}, Flava \cite{flava} and X-VLM \cite{xvlm}. For instance, although \citet{vl-bow} showed that some of these models perform reasonably well on the spatial relationships part of their VG-Relations benchmark, all the models  they evaluate (CLIP, BLIP, XVLM, Flava, NegCLIP \cite{vl-bow} and CLIP fine-tuned on COCO) discriminate between an object/person on the left versus on the right at approximately random chance (49-52\% accuracy). Given this behavior, we argue that the apparent success on other spatial relations is also not indicative of actual understanding of those relations, but rather of learning  that two particular objects or people usually co-occur with a specific, directed relation. For instance, in the VG-Relations benchmark dataset (which consists of natural images) ``a cow on the grass" is much more likely to exist in the test set than the negative caption ``grass on a cow". Figure \ref{fig:word_distribution} shows the distribution of objects/people that are most commonly ``on" objects, versus objects that are most commonly ``under" other objects in the VG-relations benchmark. This effect was quantified on the VALSE \cite{valse} benchmark: Even though VALSE explicitly tries to counter the distributional bias (``cow on grass" is much more likely than ``grass on cow") by generating hard negative sentences with roughly the same likelihood, language-only models (with no access to the image) perform better than vision-language models on VALSE's spatial relations task. NegCLIP \cite{vl-bow} improves over CLIP on all spatial relations except for ``left" and ``right", by adding negative captions with swapped words, thus making the model focus more on word order. This translates to better performance on relations with a clear common direction, but notably not in the case of ``left" versus ``right" relations.

\begin{figure}[h!]
  \begin{subfigure}{0.49\columnwidth}
    \centering
    \includegraphics[width=\textwidth,trim={0 0 1.5cm 1.5cm},clip]{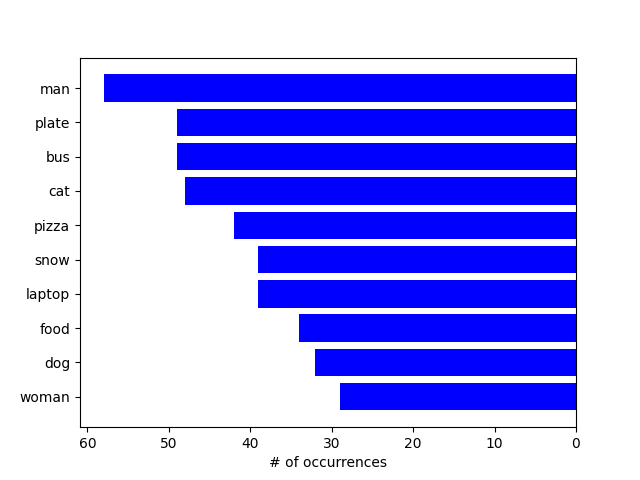}
    \caption{Mostly ``on''}
  \end{subfigure}
  \hfill
  \begin{subfigure}{0.49\columnwidth}
    \centering
    \includegraphics[width=\textwidth,trim={1.5cm 0 0 1.5cm},clip]{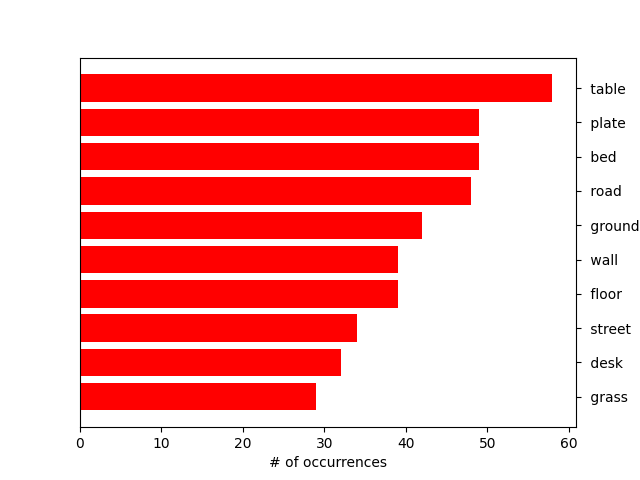}
    \caption{Mostly ``under''}
  \end{subfigure}
  \caption{Objects that are most commonly on other objects vs. objects that are usually under other objects in VG-Relations}
  \label{fig:commonobjects}
\end{figure}

\begin{figure}[h!]
    \includegraphics[width=\linewidth]{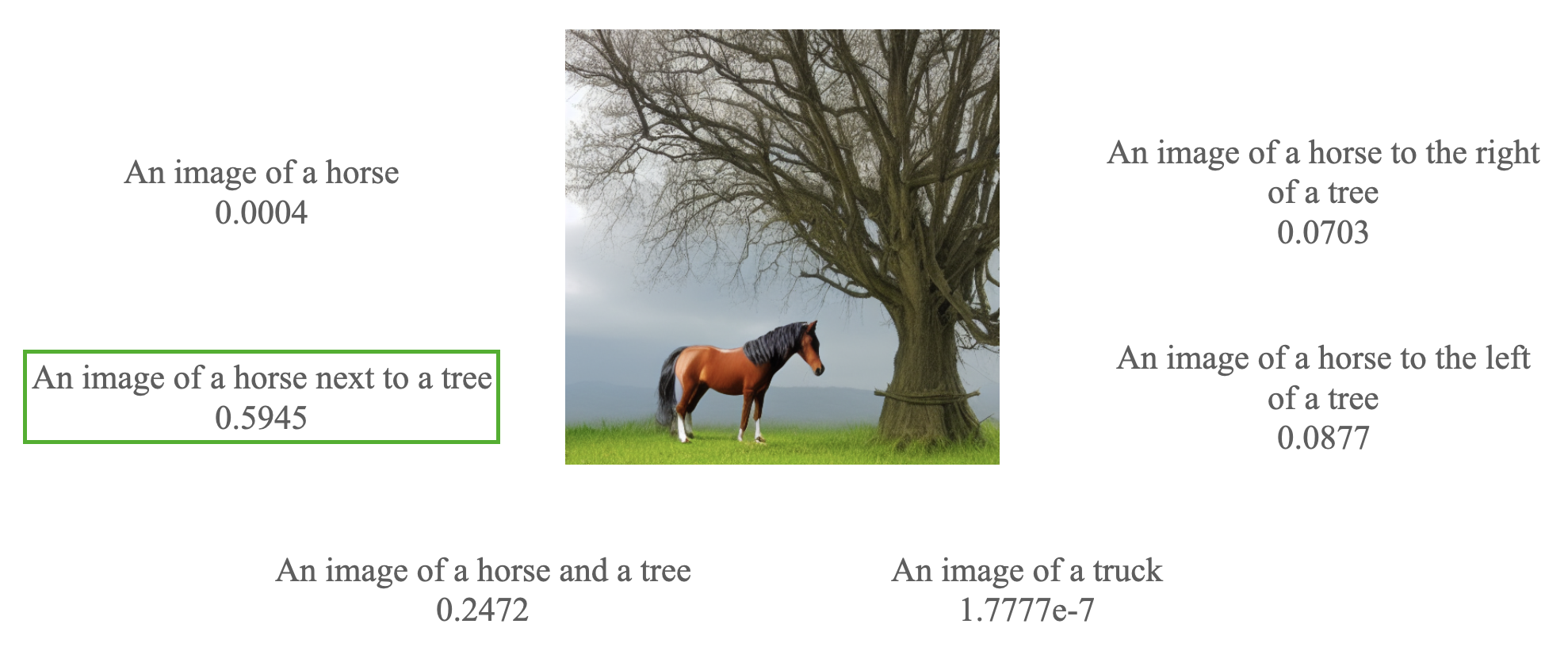}
    \caption{CLIP's perception: which text matches this image best (softmax of cosine similarity)}
  \label{fig:clip_importance}
\end{figure}

 % An example of this would be nice. 
 % Could we show a graph that maps  how common some relations are versus their inverse, ie. we invert a sentence and see does the inversion exist.
For almost all spatial relations, where $\rightarrow$ denotes a relation, clear patterns of the conventional combination exist, where either $X \rightarrow Y$ or $Y \rightarrow X$ is much more likely.  We argue that these models do not learn the abstracted meaning of the words that describe the positioning, but rather only ``succeed" because some combinations are far more likely than others in both the training and evaluation sets. This makes evaluation of the true performance of models on spacial relations challenging. For instance,  due to this imbalance in relation distribution, VG-Relations is not suited for evaluating positional understanding for most relations. We specify that this is the case in \textit{most} instances as there is a noteworthy exception to this convention -- left-right relations. These, in contrast to other relations such as ``top" and ``bottom", are balanced in their distribution, e.g. "object a to the left of object b" is equally as likely as "object a to the right of object b". Therefore, in order to accurately evaluate relative position understanding, we limit the focus of our evaluation to left-right relations.

The effects of increasing the scale of both datasets and models have also been explored. \citet{clip-scaling} show that CLIP's performance increases on ImageNet zero-shot classification and image retrieval on MS-COCO and Flickr-30k as the model size and training dataset size increases.  However, it is important to understand that these datasets do not require a lot of positional understanding  -- as evaluated by \citet{vl-bow}. When testing the capability of CLIP models to distinguish between left and right on the VG-Relations benchmark, no sign of any scaling behaviour on this task can be seen, as shown in Table \ref{scalingtable} (These results are for OpenCLIP models  and were obtained from the OpenCLIP github repository. \footnote{https://github.com/mlfoundations/open\_clip})

\begin{table}[h!]
\begin{tabular}{@{}lrr@{}}
\toprule
 \makecell[r]{CLIP\\ model}  & \makecell[r]{zero-shot accuracy\\ Imagenet 1k} & \makecell[r]{left vs. right accuracy \\  VG-Relations} \\ \midrule
ViT-B/32 & 62.9                                                                     & 50.5                                                                             \\
ViT-B/16 & 67.1                                                                     & 50.5                                                                             \\
ViT-L/14 & 72.8                                                                     & 50.0                                                                             \\
ViT-H/14 & 78.0                                                                     & 50.0                                                                             \\ \bottomrule
\end{tabular}
\caption{OpenCLIP model scaling vs performance}
\label{scalingtable}
\end{table}

\subsection{Relative Position in Embedding Space}
\begin{figure}[h!]
    \includegraphics[width=\linewidth]{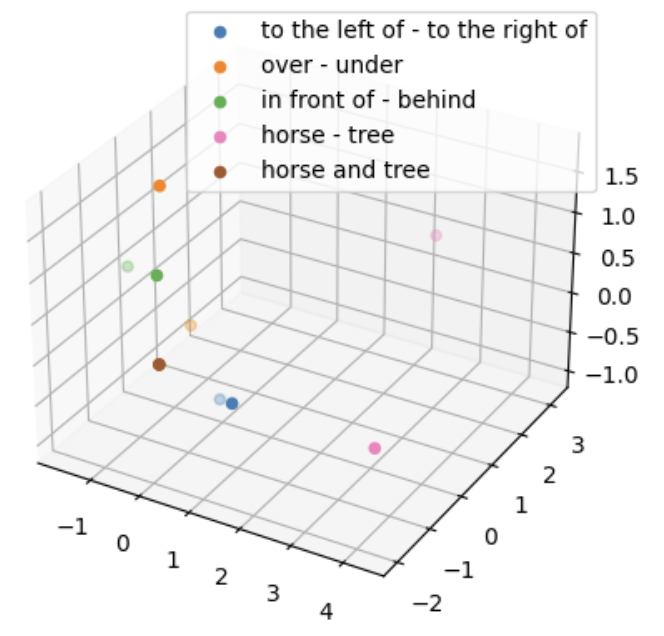}
    \caption{Proximity of spatial relations in CLIP's embedding space. Notice that most of these opposites are closer to each other than to the other points.}
  \label{fig:spatialrelations}
\end{figure}

To further interrogate the manner in which CLIP encodes relative position, we measure the distance between the two concepts ``left" and ``right" in embedding space.  Being that ``left" and ``right" are semantic opposites, we expect that they would be ``far away" from each other. However, as shown in Figure \ref{fig:spatialrelations}, when we actually measure the distance between these two concepts in embedding space, we see that they are actually very close. The reason for this counter-intuitive finding might be that current image-text datasets are not suited for teaching contrastive models this distinction and that, within  these images and text, left and right end up serving as analogues of each other.

Moreover, if we were to look at the pixel level, the difference between the two images is a simple vertical mirroring e.g. one linear transformation. Analogously,  comparing the two descriptions ``A horse is standing to the left of a tree" and ``A horse is standing to the right of a tree", on a purely textual level, they share 90 percent of the words in the same order and position. The stark semantic difference between something being on the left versus on the right of another object is therefore not inherent to the image-text data currently used to train vision-language models. 
\begin{figure}[h!]
    \includegraphics[width=\linewidth]{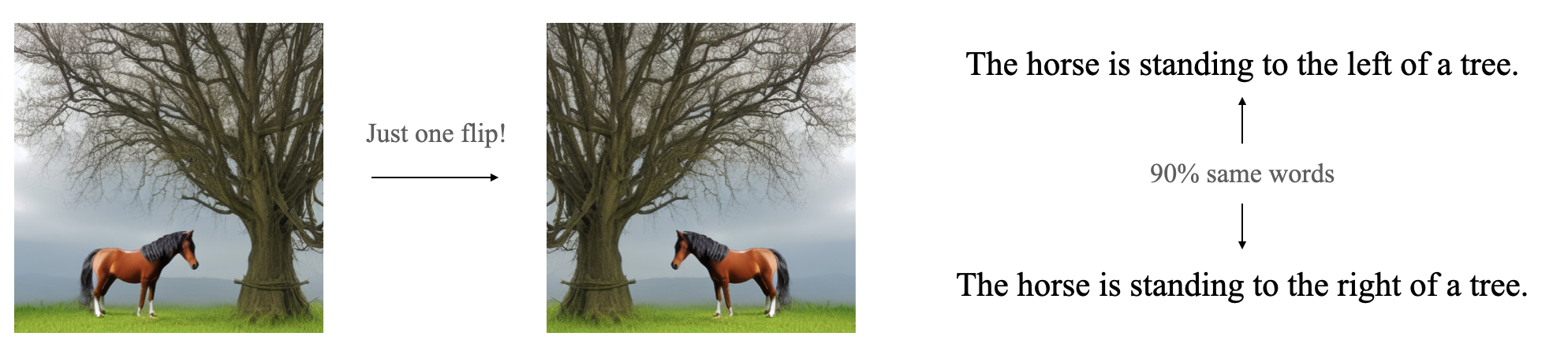}
    \caption{Example of semantically different but syntactically close image/caption pairs.}
  \label{fig:semanticsvssyntax}
\end{figure}
 
To test whether, conceptually, ``right" aand ``left" are treated similarly, we visualize the distances between opposite spatial relations and between nouns for comparison on a specific example in Figure \ref{fig:spatialrelations} (for a table of the distances please refer to Table \ref{cosine_table} in Section \ref{results}). All of the spatial relations are between a horse and a tree, so for example the vector for ``over" was obtained by encoding the sentence ``a horse over a tree". Each pair of spatial relations has one color, the first relation has a darker tone and the second has a lighter tone of the same color. We observe a significant distance between two nouns, ``horse" and ``tree" in pink, but almost no distance between ``a horse to the left of a tree" and ``a horse to the right of a tree". We also observe that ``a horse and a tree" (brown) is much closer to ``a horse under a tree" (light orange) than to ``a horse over a tree" (dark orange) indicating that the former is likely the more common occurrence.

\subsection{Positional Importance}

From an optimization standpoint, we know that it is more likely that the distinction between semantically opposite spatial concepts will be learned if this difference helps, in some way, to solve the task being learned. CLIP's contastive objective should, in theory, be able to contrast objects/people on the left with objects/people on the right, therefore learning a semantic distance between the concepts (provided that examples requiring this distinction to be learned are uniformly distributed within training batches). Given this assumption, we identify three primary reasons why this does not occur: 
\begin{enumerate}
    \item The number of captions containing the words ``left" and ``right" is minimal in COCO, as illustrated in Figure \ref{fig:word_distribution}
    \item To adequately learn the distinction between ``left" and ``right", the same two objects with the opposite relation must be present together in one specific mini-batch. Otherwise, the model will likely be able to pick up on more superficial auxiliary information and be able to disregard the spatial locations entirely. Choosing the image with the caption ``a tiger on the left side" is easy if there is only one image of a tiger in the mini-batch. The model will therefore not learn to distinguish between ``a tiger on the left side" and "a tiger on the right side" as it has not been ``forced" to do so.
    
    \item The relative position stated within the caption relies on the interpretation of the annotator. It is entirely possible for the annotator of a given example to invert the left-right relation to align with the viewpoint of the object/person the relation refers to i.e. the notion of ``my left versus your left". In particular ``left" and ``right"  tend to vary far more than other relations such as ``top" and ``bottom" in this regard due to this frequent inversion. Without manual correction of the viewpoint within large-scale datasets, we cannot determine whether left and right have been inverted.
\end{enumerate}

The observations we've made so far seem to indicate that CLIP does not assign a high degree of importance to some positional information. In fact, we actually observe this in the outputs of the model. In some cases, CLIP assigns a lower probability to sentences containing spatial information than to more vague sentences relating objects with conjunctions like ``and" or less precise prepositions like ``next to". An example of this behaviour can be seen in Figure \ref{fig:clip_importance} where a caption containing ``next to" is considered closer to the image than ``to the left of".  
When looking at the general similarities between embeddings of images and captions this does not seem to hold. Table \ref{templatesimilarities} displays the mean cosine similarities between each of the captions and an image. The images were selected from VG-Relations such that object A is to the left of B. We do not include images where object A is to the right of B so that they don't cancel each other out in the averaged similarities. These distances indicate that, in general,  captions containing ``to the left of" and ``to the right of" are seen as even slightly more likely by CLIP than those containing conjunctions like ``and" or less precise language like ``next to". 
The same trend is also visible in our synthetic dataset, which will be described in Section \ref{synthdata}. However since only about one in 500 captions in COCO contains either the word ``left" or ``right", just slight trends can skew the probability distribution learned by CLIP, leading it to assign a higher probability to imprecise language in some constellations as shown in Figure \ref{fig:clip_importance}.

\begin{table}[h!]
\begin{tabular}{@{}lrr@{}}
\toprule
         caption template  & similarity \\ \midrule
An image of A & 0.227384                                                                                     \\
An image of A next to B & 0.240967                                                                                             \\
An image of A and B & 0.246005                                                                                                   \\
An image of A to the left of B & 0.252456                                                                                                    \\
\bottomrule
\end{tabular}
\caption{Mean cosine similarities between CLIP embeddings of images and captions}
\label{templatesimilarities}
\end{table}

\section{Learning Relative Position}
\label{experimental-setup}
 In this section, we propose a simple strategy to address the shortcomings discussed in Section \ref{background}. We construct a synthetic dataset that more carefully encodes specific positional relations. We fine-tune CLIP using our synthetic data and provide a discussion on the performance.

\subsection{Synthetic Data Generation}
\label{synthdata}

\begin{figure}[h!]
  \begin{subfigure}{0.49\columnwidth}
    \centering
    \includegraphics[width=\textwidth]{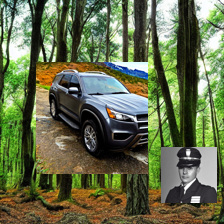}
      \caption{"A car to the left of a policeman"}
  \end{subfigure}
  \hfill
  \begin{subfigure}{0.49\columnwidth}
    \centering
    \includegraphics[width=\textwidth]{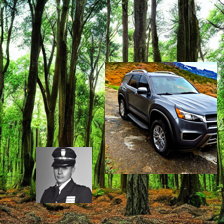}
    \caption{"A policeman to the left of a car"}
  \end{subfigure}
  \caption{A contrastive image-text pair}
  \label{fig:minibatch}
\end{figure}

We construct a synthetic training dataset by extracting all nouns from VG-Relations using SpaCy \cite{spacy}. For each of these 867 nouns, we generate several images using Stable Diffusion 1.4 via Hugging Face \cite{huggingface}. We then sample random pairs of objects $A$ and $B$, randomly resize both objects and select one out of a diverse set of generated backgrounds. 
We create two images that only differ in the positioning of the objects: one where $A \rightarrow B$ ($\rightarrow$ denotes "to the left of") and one where $B \rightarrow A$. Captions for both are generated accordingly. Since $A \rightarrow B$ means the same as $B \leftarrow A$ we use both of those synonyms in 50\% of captions.
Each of these contrastive pairs, consisting of two images and their corresponding captions as shown in \ref{fig:minibatch}, will be included in the same mini-batch during training. (as two independent image-text pairs) This means that to optimize the contrastive loss the model is forced to acquire positional understanding, because the images only differ in the positioning of the objects and the captions only differ in the direction of the relation between A and B.

More specifically, the data generation process works as follows:
\begin{enumerate}
  \item Extract all nouns from VG-Relations
  \item Generate 10 images for each noun using Stable Diffusion (Since the Hugging Face Stable Diffusion pipeline automatically detects and censors NSFW generations, we keep generating until 10 usable images are produced)
  \item Generate a diverse set of backgrounds from prompts like "a kitchen", "a forest", "a beach" etc. 
\end{enumerate}

Then, to stitch synthetic images together, we: 
\begin{enumerate}
  \item Select a random background
  \item Sample a random pair of nouns (without replacement). \item For each noun, select one of the 10 images
  \item Resize both images randomly
  \item Place the first image on the background randomly, but leaving at least enough space that the second image can still satisfy the spatial relation
  \item Place the second image randomly, so that it satisfies the spatial relation
\end{enumerate}

\subsection{Evaluation Dataset}

The pre-requisite for an effective evaluation dataset for this task is that the dataset contains a sufficient number of examples displaying spacial relationships. Winoground \cite{winoground}, VALSE \cite{valse} and Visual Genome Relations \cite{vl-bow} test on realistic scenes (although Winoground uses generated images) and pose the task as triplets consisting of an image, a positive caption and a negative caption. This format allows for evaluation with embedding models by comparing the probability of the image matching the positive caption as opposed to  the negative caption. The goal is to maximize the pairwise ranking accuracy, i.e. the prediction is correct if, for an image $x_i$ ,   $p(x_i, c^{True}_i) > p(x_i, c^{False}_i) $. Winoground however is very small and covers a wide range of compositional phenomena, commonsense reasoning and other niche abilities \cite{whywinohard} It's broad scope means that it contains very few examples that specifically target positioning in images. VALSE also only contains 535 examples concerning  spatial relations. Another related dataset is the Visual Spatial Reasoning (VSR) \cite{vsr} benchmark to classify captions as true or false. This, however, does not allow for zero-shot evaluation of CLIP-like embedding models as there is no negative caption to compare to. 
%Models are tested after fine-tuning on the VSR training set. 
What is left and what is right also depends on the perspective. VSR for example has a lot of categories that require perspective change and is thus much more complicated.

For this evaluation,  we performed zero-shot evaluation, as we are interested in testing for existing knowledge and to minimize the relationship between the training and evaluation dataset, since that can significantly bias the embedding space to the target task \cite{finetuning-embeddingspace}. The annotations for the Visual Genome (VG) Relations \cite{vl-bow} are constructed from scene graphs so ``left" and ``right" consistently refer to the viewer's perspective. Additionally,  we want to test on the most diverse set of objects making VG-Relations even more appropriate for evaluation, since a large proportion of the dataset tests for spatial understanding on natural images. Due to the evaluation issues with most spatial relations outlined in section \ref{background}, however, we conclude that the spatial relationships section of VG-Relations is only suited for evaluation of left versus right. We deem it worthy to focus on this constrained area because even the best model at this task, XVLM \cite{xvlm} performs at only 52 percent accuracy on this binary classification task. 

 To evaluate this specific relation, we use all samples containing ``left" and ``right" from the spatial relationships subtask of the VG-RelationS benchmark leaving us with a total of 15,482 samples. Each of the samples consists of an image, a positive caption and a negative caption. We then calculate the cosine similarity of the image embedding with each of the captions to determine the caption that fits best and report the accuracy.

\subsection{Training}
Except for the scaling comparison, OpenAI's CLIP ViT-B-32 was used in all cases to enable a fair comparison to NegCLIP, which also uses the ViT-B-32 architecture. Text inputs were preprocessed with OpenCLIP's byte-pair-encoding tokenizer and images were resized to $224\times224$ pixels and normalized via OpenCLIP's image transform. We fine-tune a CLIP ViT-B-32 model on our synthetic data using the OpenCLIP implementation \cite{openclip} for 15 epochs. We use the following sampling routine: Each batch of size 512 contains a share of randomly sampled image-text pairs from MSCOCO \cite{coco} and a smaller number of contrastive pairs (the pairs of two images and their captions as depicted in figure \ref{fig:minibatch}) out of our synthetic dataset, we'll denote this as $cs$.
This ensures that for the $cs$ share of the mini-batch retrieving an image from a caption requires differentiating between two very similar images and vice versa.
This is similar to how \citet{vl-bow} augment their text data to create hard negatives. However our alternative images are much stronger alternatives since they are specially constructed for this purpose whereas  \citet{vl-bow} sample semantically similar alternative images, and their captions from COCO, which yields much less relevant images. The model was trained on a Tesla V100 32GB GPU and hyper-parameters are summarised in Table \ref{hyperparams}.
\subsection{Evaluation}

We take a similar experimental approach to \citet{vl-bow}, who compare the collowing models: CLIP \cite{clip}, BLIP \cite{blip}, Flava \cite{flava}, X-VLM \cite{xvlm} and their own NegCLIP \cite{vl-bow}.

\begin{table}[h!]
\centering
\caption{Hyper-parameters for CLIP ViT-B-32, \textit{cs} denotes the number of contrastive samples within one mini-batch. Since they go in pairs, the number of contrastive pairs is given by  $cs/2$} \label{hyperparams}
\begin{tabular}{@{}lcccr@{}}
\toprule
 \begin{tabular}[c]{@{}c@{}}learning\\rate\end{tabular}& \begin{tabular}[c]{@{}c@{}}weight\\decay\end{tabular} & \begin{tabular}[c]{@{}c@{}}warm\\-up\end{tabular} & \begin{tabular}[c]{@{}c@{}}batch\\size\end{tabular}& \begin{tabular}[c]{@{}c@{}}cs\end{tabular}  \\ \midrule
1e-4 & 0.2 & 2000 & 512 & 64 \\ \bottomrule
\end{tabular}

\end{table}
We use Weights and Biases for logging \cite{wandb}.
\section{Results}
\label{results}

Table \ref{left-right results} summarises the results of our experiments. Results are obtained using the ARO implementation \footnote{https://github.com/mertyg/vision-language-models-are-bows}. We can see that we achieve, for the first time to our knowledge, an accuracy score significantly over random chance on the distinction between ``left" and ``right". This demonstrates that training relational understanding on stitched synthetic images manages to generalise to natural images.

\begin{table}[h!]
\centering
\caption{Accuracy on the distinction of ``to the left of" vs ``to the right of" on the VG-Relations benchmark}
\label{left-right results}
\begin{tabular}{@{}lrr@{}}
\toprule
\multirow{2}{*}{Model} & \multicolumn{2}{c}{Accuracy}\\ \cmidrule{2-3}
& ``left of" & ``right of" \\ \midrule
CLIP     & 0.49              & 0.49               \\
NegCLIP  & 0.50              & 0.50               \\
CLIP-FT  & 0.50              & 0.50               \\
XVLM     & 0.52              & 0.52               \\
BLIP     & 0.51              & 0.49               \\
Flava    & 0.50              & 0.51               \\
\makecell[l]{OURS}     & 0.57              & 0.56    
\\
\bottomrule
\end{tabular}
\end{table}

\begin{figure*}[ht]
  \centering
  \begin{minipage}[b]{0.24\textwidth}
    \centering
    \includegraphics[width=4cm]{images/vgrelations/2414605.jpg}
    \caption{``the wall is to the left of the horse"}
  \end{minipage}
  \hfill
  \begin{minipage}[b]{0.24\textwidth}
    \centering
    \includegraphics[width=3.5cm]{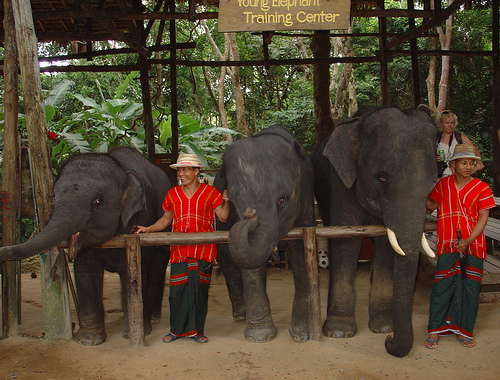}
    \caption{``the fence is to the right of the elephant"}
  \end{minipage}
  \hfill
  \begin{minipage}[b]{0.24\textwidth}
    \centering
    \includegraphics[width=4cm]{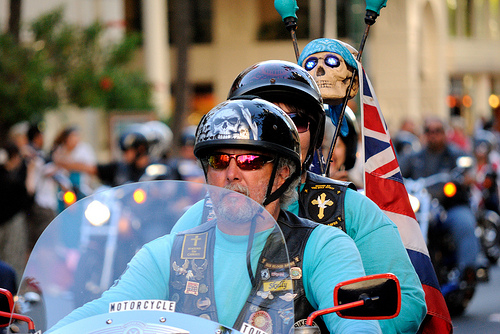}
    \caption{``the woman is to the right of the tree"}
  \end{minipage}
  \hfill
  \begin{minipage}[b]{0.24\textwidth}
    \centering
    \includegraphics[height=2.7cm]{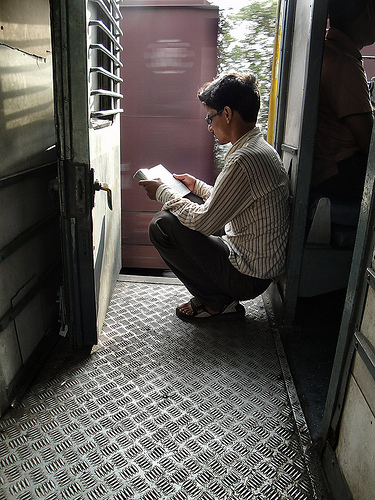}
    \caption{``the shirt is to the right of the door"}
  \end{minipage}

  \caption{VG-Relations contains ambiguous/nonsensical image-caption pairs.}
  \label{fig:abiguous}
\end{figure*}

An important caveat to these evaluation results is that the captions in VG-Relations are automatically generated from scene graphs. As a result of this, a considerable portion of the captions are very ambiguous, making this task challenging at times -- even for humans. For reference, we included some ambiguous samples from VG-Relations in Figure \ref{fig:abiguous}

We now look to the embedding space to evaluate whether the distinction between ``left" and ``right" is reflected within it. Table \ref{cosine_table} shows that while for CLIP and even NegCLIP both concepts are very close to each other, our model spaces them much further apart, indicating a stronger understanding that ``left" and ``right" are semantic opposites. We compare the distance between the concepts ``left" and ``right" to the distance between two random nouns as a reference point for how large distances in the model's embedding space generally are. We use 866 pairs of nouns out of the 867 different nouns present in VG-Relations.

\begin{figure}[h!]
    \includegraphics[width=\linewidth]{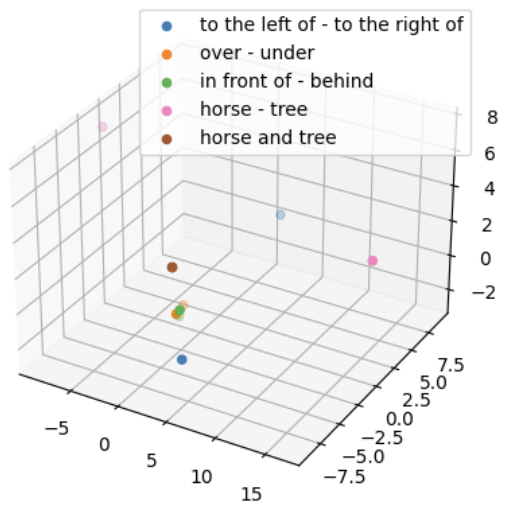}
    \caption{Proximity of spatial relations in our model's embedding space}
  \label{fig:spatialrelationsresults}
\end{figure}

We can also observe the distance between ``left" and ``right" created by our model in the embedding space plot in Figure \ref{fig:spatialrelationsresults}. Our model keeps a significant distance between ``horse" and ``tree", but also clearly separates ``left" and ``right". Also ``horse and tree" lie almost in-between ``left" and ``right" (in contrast to Figure \ref{fig:spatialrelations}), which aligns well with the semantics.
Since we did not train to distinguish between ``over", ``under", ``in front of" and ``behind", those are still placed very close to each other.

\begin{table}[h!]
\centering
\caption{Cosine similarites between the concept of left and right and the cosine similarities between two random nouns for comparison. ``X to the right of Y” and ``X to the left of Y” are denoted by $X \rightarrow Y$ and $X \leftarrow Y$ respectively.
X and Y are randomly assigned from the nouns occurring in VGR. The similarites are then averaged.}\label{cosine_table}
\setlength{\tabcolsep}{2pt}
\begin{tabular}{@{}lrr@{}}
\toprule
Model   & \makecell{$s(X \rightarrow Y ,$\\$ X \leftarrow Y)$} & $s(X, Y)$                     \\ \midrule
CLIP    & $0.9894\pm0.0038$         & $0.7901 \pm 0.0462$ \\
NegCLIP & $0.9868\pm0.0037$         & $0.8309 \pm 0.0405$ \\
\makecell[l]{OURS} & $0.5947\pm0.0905$         & $0.1732 \pm 0.1244$ \\
\bottomrule
\end{tabular}
\end{table}

\subsection{Discussion}
In recent years, a trend has emerged where models are built on a larger and larger scale and we hope for ``emergent" capabilities. However, as we have shown, some skills like positional understanding in contrastive vision-language models like CLIP don't seem to be solvable by scaling. We would like to point out that there might be many more such areas where focus on a very specific problem is required to advance. There may also be some areas where, analogously to what we found about the lack of suitable benchmarks for positional understanding, sensible benchmarks are missing.

As CLIP's contrastive loss optimizes for retrieval from each mini-batch, it should not be surprising that CLIP does not learn much positional information by default, since discerning between a set of random images typically does not require the use of positional information, because they are different in many more obvious ways. If it is enough to regard “left” and ``right” as stop words to solve the task, then the models will do so. After a phase of performance boosts through model and dataset scaling there appear to be some topics that are intractable through scaling and may require more manual attention. We would like to motivate the construction of further training datasets as well as benchmarks for underdeveloped capabilites in contrastive vision-language models.

\subsection{Limitations}
Since our training dataset was generated fully automatically, some of the images do not fit their description well. Manual correction of labels could help address this.  Also, since we focussed on the distinction between left and right there is exciting room for expansion to other spatial relations in future work. 
Furthermore, in future we plan on extending these experiments to show the effects of placing opposing relations in different configurations within training batches. 
\section{Conclusion}
\label{conclusion}
We presented a simple, yet effective approach for teaching CLIP to learn left-right positional relations through the use of contrastive pairs. Our approach, for the first time, enables CLIP to learn the distinction between ``left" and ``right"  over random chance on the VG-Relations benchmark. We show that this method also better aligns the model's embedding space to human perception, placing the semantically opposite concepts of ``left" and ``right" further apart. We hope to extend this analysis to a wider set of relations and generate similarly effective contrastive pairs.

\section*{Acknowledgements}
\label{acknowledgements}

We would like to thank Thomas Laurent, Eduard Bartolovic, Julie Berndsen, Derek Greene and Tony Veale for their helpful comments on earlier versions of the paper. We also thank the anonymous reviewers for their valuable feedback.

This publication has emanated from research conducted with the financial support of Science Foundation Ireland under Grant number 18/CRT/6183. For the purpose of Open Access, the author has applied a CC BY public copyright licence to any Author Accepted Manuscript version arising from this submission. This work was supported, in part, by Science Foundation Ireland grant 13/RC/2094\_P2.

\bibliography{bibliography}
\bibliographystyle{icml2023}

\newpage
\appendix
\onecolumn

\end{document}